# Evaluation of Architectural Synthesis Using Generative AI

*A case study on Palladio's architecture*

Jingfei Huang[1] and Alexandros Haridis[2]
[1,2]*Harvard University.*
[1]*jingfeihuang@mde.harvard.edu, 0009-0002-0213-4160*
[2]*haridis@seas.harvard.edu, 0000-0003-0369-1428*

**Abstract.** Recent advancements in multimodal Generative AI may democratize specialized architectural tasks like interpreting technical drawings and creating 3D CAD models which traditionally require expert knowledge. This paper presents a comparative evaluation study of two systems—GPT-4o and Claude 3.5—in the task of architectural 3D synthesis. It takes as a case study two buildings in Palladio's *Four Books of Architecture* (1965): Villa Rotonda and Palazzo Porto. High-level architectural models and drawings of the buildings were prepared inspired by Palladio's original text and drawing corpus. Through sequential text and image prompting, the study characterizes intrinsic abilities of the systems in (1) interpreting 2D/3D representations of buildings from drawings, (2) encoding the buildings into a CAD software script, and (3) self-improving based on outputs. While both systems successfully generate individual parts, they struggle to accurately assemble these parts into the desired spatial relationships, with Claude 3.5 showing overall better performance, especially in self-correcting its output. The study contributes to ongoing research on benchmarking the strengths and weaknesses of off-the-shelf AI systems in intelligent human tasks requiring discipline-specific knowledge. The results show the potential of language-enabled AI systems to act as collaborative technical assistants in the architectural design process.

## 1. Introduction

Architectural design has traditionally been a human-centric process, largely relying on architects' creativity, intuition, and technical knowledge. This is expressed, for example, through the various representational media used in the architectural design process, such as text, symbol notations, line drawings, visualizations, and physical models. While computational design and generative algorithms expand the possibilities for exploring design ideas beyond the limitations of traditional methods, specialized tasks such as 3D CAD modelling and design scripting still demand proficiency with digital computing and software tools potentially excluding in this way a broader range of non-technical contributors to the design process.

Recent advances in Artificial Intelligence (AI) technology have resulted in a substantial increase in interest and effort toward the mechanisation of creative processes across design, art, science, and technology (Haridis, 2024). In particular, Generative AI technologies powered by Large Language Models (LLMs), Latent Diffusion Models (LDMs), and Large Vision Models (LVMs), demonstrate significant capabilities in interpreting and generating both language and image-based content. In architecture and design, current studies explore the capabilities and limitations of human-AI collaboration in creating architectural visualizations and surveys (Cobb, 2023; Ploennigs & Berger, 2023), generating plans (e.g., Babakhani, 2023), computing and visualizing structural performance (Danhaive & Mueller, 2021), comprehending engineering requirements in technical documentations (Doris et al., 2024) or generating CAD models for digital design and manufacturing (Chong et al., 2024; Edwards et al., 2024; Makatura et al., 2023; Picard et al., 2024). This recent stream of evaluative studies suggests a need for understanding whether Generative AI technologies can potentially democratise specialised technical tasks in architecture and design. Such democratization can help expand access to design tools beyond trained professionals and to empower a wider audience to foster greater inclusion and diversity within the field (the potential impact of Generative AI on accessibility of technology is discussed widely in domains beyond architecture, too; see, for example, in Epstein et al (2023)). However, the comparative effectiveness of such Generative AI systems in architectural applications, particularly in interpreting architectural documents and generating 3D CAD models of buildings, remains largely unexplored.

This paper presents a structured comparative evaluation of two off-the-shelf multimodal Generative AI systems, GPT4o (OpenAI, 2024) and Claude Sonnet 3.5 (Anthropic, 2024), in the task of architectural 3D synthesis. It takes as a case study two buildings from Andrea Palladio's treatise *The Four Books of Architecture* (1965): Villa Rotonda for its symmetrical villa design with a circular organization of architectural elements (e.g., loggias); and Palazzo Porto for its linear arrangement involving three distinct building masses. Palladio's work provides an ideal test case due to its comprehensive documentation of design principles and detailed architectural descriptions. New high-level architectural drawings and 3D models of the buildings were prepared, including sections, plans, axonometric views and part-to-whole relational diagrams. These were based on Palladio's original texts and images which include detailed descriptions of design elements and spatial organisation. Using these prepared visual material as input, a sequential text and image prompting methodology is designed to evaluate the intrinsic abilities of the two selected systems in: (1) interpreting the 2D/3D drawings of the buildings, (2) encoding the buildings into CAD scripts using the OpenSCAD software platform (Kintel & Wolf, 2021), and (3) self-improving based on their outputs. Elements of this methodology are inspired by related research at the intersection of design, engineering, and machine learning (e.g., Picard et al., 2024). However, the evaluative study presented here is novel in its use of 3D models with the spatial complexity that is characteristic of a mature architectural design language.

The results indicate that both systems are successful in encoding individual building parts, but both struggle to accurately assemble these parts into the desired spatial relationships with Claude 3.5 showing overall better performance, especially in self-correcting its output. The results showcase the capabilities and limitations of two leading Generative AI systems in handling complex spatial information and architectural details, which are often not found in other design areas where these systems are employed, such as engineering products or graphic design. The study thus contributes to current discussions across disciplines in benchmarking the strengths and weaknesses of off-the-shelf AI systems in human tasks requiring discipline-specific knowledge. The results of this paper provide insights into the potential of language-driven AI systems to become collaborative technical assistants in the architectural design process. The original drawings, prompt design, and methodological details are provided in a publicly accessible repository (Huang & Haridis, 2025).

## 2. Methodology

This study follows the evaluation pipeline described in the diagram in Figure 1. Broadly speaking, the evaluation methodology involves: (a) corpus selection and preparation, (b) prompt design and interpretation of architectural drawings, and (c) assessment of AI system performance in CAD generation and self-correction using a matrix scoring system. These three aspects of the evaluation methodology are described in more detail in the following sections. The generative AI systems used here are GPT-4o in the ChatGPT Playground (version of May 13, 2024) and Claude 3.5 Sonnet (version of October 22, 2024) in the Anthropic Workbench.

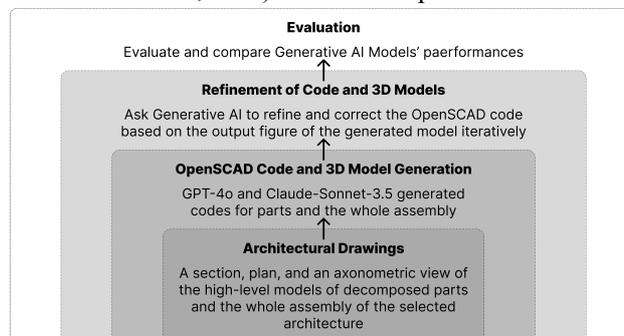

*Figure 1. Study Pipeline*

## 2.1. CASE SELECTION AND CORPUS PREPARATION

This paper uses as case study two buildings from Andrea Palladio's treatise *The Four Books of Architecture* (Palladio, 1965). The following two buildings were selected because they represent distinct spatial organizational principles:

- **Villa Rotonda**: This work was selected for its symmetrical single-volume composition, featuring a central dome with four identical side halls and loggias. The building was decomposed into three parts: the four entrances (the outermost elements, consisting of the loggias), the side hall (the middle mass connecting the entrances to the central hall), and the central hall (the core domed space).

- **Palazzo Porto**: This building was chosen for its more linear spatial arrangement, which demonstrates a non-circular spatial relationship. This building is decomposed into three main parts: the main buildings (two masses at both ends of the building containing the entrance halls and rooms), the courtyard (the central space located between the main buildings), and the primary staircase connected to the courtyard.

After obtaining unsatisfactory results from a preliminary pilot study that used Palladio's own detailed drawings, new custom high-level models were created for the selected buildings to serve as an input to the two AI systems (see Figure 2). The abstract modelling approach followed in this study is related to other approaches in the literature for the representation of high-level spatial architectural information, such as 3D massing models (e.g., Arslan et al., 2022), or two-dimensional abstracted plans as in studies of villa plans in the shape grammar framework (Stiny & Mitchell, 1978). Digital 3D models as well as plans, sections, axonometric views, and part-to-whole relational diagrams were all prepared using the Rhino 8 software platform (McNeel & Associates, 2024). This visual-spatial material was then organised into a single drawing sheet (see Figure 3) and used as an input prompt for the AI systems as described below.

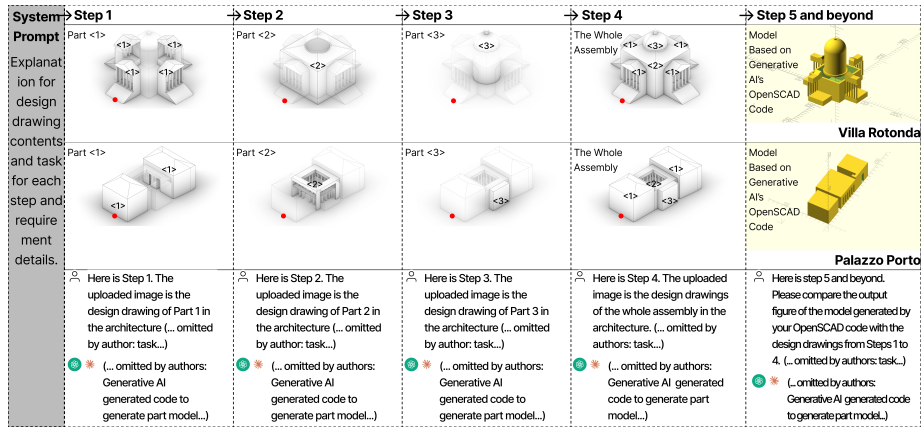

*Figure 2. Study Process Diagram of The Interaction Steps: system prompt, steps 1 to 3 of prompting parts, step 4 of prompting the whole assembly, and step 5 and beyond for the iterative refinement*

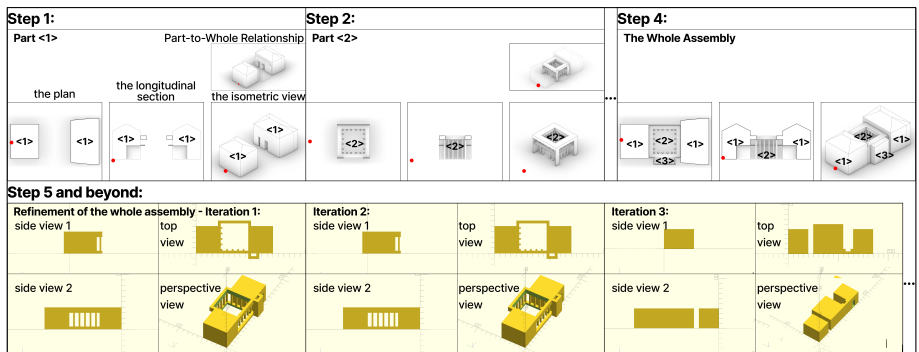

*Figure 3. Examples of Drawing Sheets Used for Prompting in Step 1 (top left), Step 2 (top middle), Step 4 (top right) and Step 5 iterations (bottom) for the case of Palazzo Porto*

## 2.2. PROMPT DESIGN

As Figure 2 shows, a step-by-step prompting process was carried out for each separate model, ensuring that conversation histories were not shared between sessions. Both systems were configured with a temperature setting of 1.0 and a specified maximum token limit. The prompts were structured as follows:

- **System Prompt (or system instruction):** The AI's role was explicitly stated as "an expert proficient in OpenSCAD scripting and architectural design." The system was informed that it would receive images documenting a building and was provided with detailed instructions for tasks from Step 1 to Step 5. The system's goal was to generate OpenSCAD code that generates a 3D model that matches the design drawings. Specific requirements for the task included utilising the red dot as the origin in the shared coordinate system for all parts, and ensuring an accurate representation of dimensions, positions, orientations, and alignments of parts on a shared XY plane.

- **Step 1 to 3**: The system is provided with prepared drawings for individual parts. These include a plan, a section, an axonometric view, a part-to-whole diagram, and the textual prompts with step number, task, and requirements.

- **Step 4**: The system is provided with prepared drawings for the whole building, including a plan, a section, an axonometric view, and the textual prompts to indicate the step number, task, and requirements.

- **Step 5 and beyond**: The system is provided with prepared drawings showing four views of the model generated by the OpenSCAD code in Step 4, with the prompts indicating the step number, task, and requirements. Then, Step 5 was repeated as necessary until satisfactory results were achieved or the outputs became repetitive or began to deteriorate, indicating the AI system has reached its capacity for self-refinement.

## 2.3. EVALUATION FRAMEWORK

To evaluate model performance, this study adopted the method described by Picard et al. (2024) for assessing the performance of a vision-language model in interpreting engineering design drawings and CAD generation. In particular, a matrix in a binary scoring system is designed with three components:

- One point is given for successfully mentioning all parts.

- The average of three points is given for the successful generation of each part in the model: one point for placing the part in the model, one point for generating the part in the correct proportion, and one point for correctly orienting the part.

- The success of the overall building generation is calculated by accumulating the following points:

    o One point for correctly positioning each part.

    o One point for correctly orienting the building.

    o One point for generating the building in the correct proportions.

      ○    One point for generating correct spatial relationship between parts.

Scores from each part were summed to provide a quantitative measure for a systematic comparison of the two AI systems' performance.

## 3. Result

### 3.1. THE CASE OF VILLA ROTONDA

#### 3.1.1. GPT 4o Result

Model results and evaluations for GPT-4o in the case of Villa Rotonda are given in Figure 4. From Step 1 to Step 3, the system successfully generates the Main Hall and Side Hall parts, but it fails to generate the four loggias with the correct proportions and orientations. In Step 4, the system assembled all the parts but did not self-correct proportional inaccuracies of the Entrance part or incorrect spatial relationships between the rest of the components. From Step 5, the system self-improves for certain part details, but the overall proportions and spatial relationships are still missing. GPT-4o achieved 61.56% average performance, with limitations in generating correct proportion and spatial relationships.

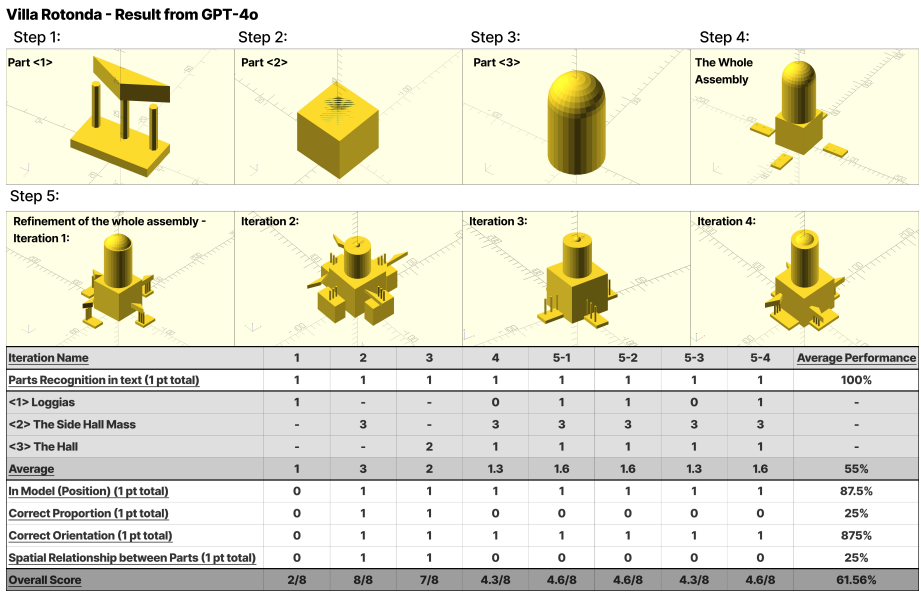

| Iteration Name | 1 | 2 | 3 | 4 | 5-1 | 5-2 | 5-3 | 5-4 | Average Performance |
|---|---|---|---|---|---|---|---|---|---|
| Parts Recognition in text (1 pt total) | 1 | 1 | 1 | 1 | 1 | 1 | 1 | 1 | 100% |
| <1> Loggias | 1 | - | - | 0 | 1 | 1 | 0 | 1 | - |
| <2> The Side Hall Mass | - | 3 | - | 3 | 3 | 3 | 3 | 3 | - |
| <3> The Hall | - | - | 2 | 1 | 1 | 1 | 1 | 1 | - |
| Average | 1 | 3 | 2 | 1.3 | 1.6 | 1.6 | 1.3 | 1.6 | 55% |
| In Model (Position) (1 pt total) | 0 | 1 | 1 | 1 | 1 | 1 | 1 | 1 | 87.5% |
| Correct Proportion (1 pt total) | 0 | 1 | 1 | 0 | 0 | 0 | 0 | 0 | 25% |
| Correct Orientation (1 pt total) | 0 | 1 | 1 | 1 | 1 | 1 | 1 | 1 | 875% |
| Spatial Relationship between Parts (1 pt total) | 0 | 1 | 1 | 0 | 0 | 0 | 0 | 0 | 25% |
| Overall Score | 2/8 | 8/8 | 7/8 | 4.3/8 | 4.6/8 | 4.6/8 | 4.3/8 | 4.6/8 | 61.56% |

*Figure 4. The Result Images of Villa Rotonda from GPT4o from Step 1 to Step 5 (top), and The Evaluation Matrix Result (bottom)*

#### 3.1.2. Claude 3.5 Sonnet Result

Model results and evaluations for Claude 3.5 in the case of Villa Rotonda are given in Figure 5. From Steps 1 to 3, the system generated individual parts with correct positions but failed to correctly locate the four components in the Entrance. From Step 4 onward, the model's generated assembly of parts showed inaccuracies in proportions and spatial relationships between parts. Despite these, the system demonstrated consistent refinement of details in individual building parts. Claude 3.5 Sonnet achieved an average performance of 73.12%, outperforming GPT-4o in generating detailed parts and self-correcting, but it faces similar challenges in maintaining correct spatial relationships.

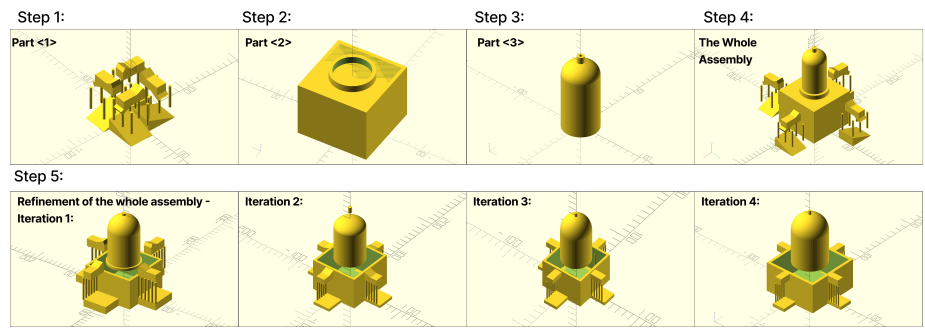

*Figure 5. The Result Images of Villa Rotonda from Claude 3.5 Sonnet from Step 1 to Step 5 (The evaluation tables were omitted due to the space limitation. See the full evaluation table at the repository (Huang & Haridis, 2025))*

## 3.2. THE CASE OF PALAZZO PORTO

### 3.2.1. GPT 4o Result

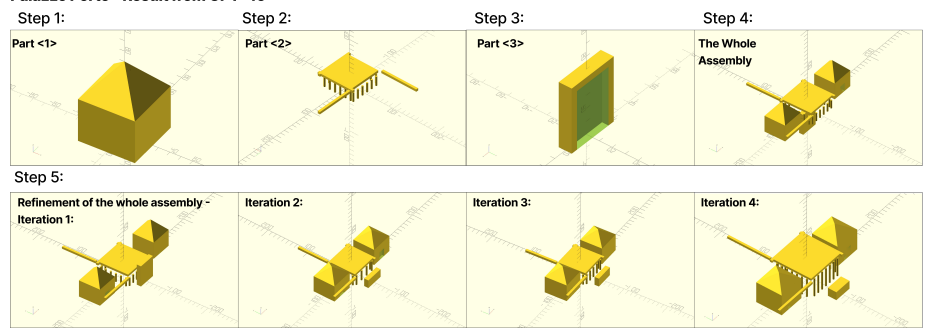

*Figure 6. The Result Images of Palazzo Porto from GPT4o from Step 1 to Step 5 (The evaluation tables were omitted because of the limitation of space. See the full evaluation table at the repository (Huang & Haridis, 2025))*

Model results for GPT-4o and Claude 3.5 in the case of Palazzo Porto are given in Figure 6. From Steps 1 to 3, individual building parts were generated with correct positions and orientations, though the proportions of the Main Building and Courtyard parts were incorrect. In Step 4, the assembly displayed coherence, accurately recreating the position, orientation, and spatial relationships between parts. However, errors persisted in generating the correct proportions for the Courtyard and Staircase parts. From Step 5 onward, iterative refinements improved the proportions of the Staircase part and added details to the Main Building part (Figure 6). Despite these refinements, there is still room for improving the overall proportions. GPT-4o achieved an average performance of 65%, with weaknesses in correcting spatial relationships, especially during the iterative self-improvement stages.

### 3.2.2. Claude 3.5 Sonnet Result
From Steps 1 to 3, all parts were generated with correct positions, proportions, and orientations. In Step 4, the assembly appeared visually consistent with the design drawings, except for an incorrect spatial relationship of the Stair part relative to the rest of the model. From Step 5 onward, refinements focused on fixing the spatial relationship through reorientation and repositioning but were unsuccessful (see the Step 5 result in Figure 7). Additional details were added to the Main Building part during the final iterations. Claude 3.5 Sonnet achieved an average performance of 73.43%, earning higher scores for correctly representing individual parts, the overall position, proportion with lower scores in maintaining correct orientation compared to results from GPT-4o. The complete evaluation tables are available in the public repository (Huang & Haridis, 2025).

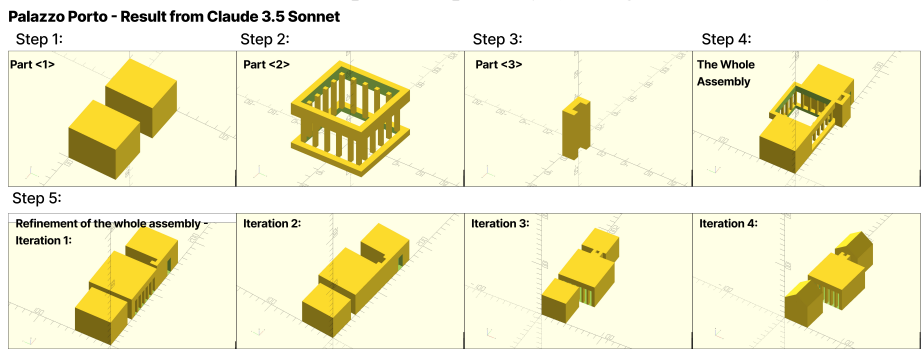

*Figure 7. The Result Images of Palazzo Porto from Claude 3.5 Sonnet from Step 1 to Step 5 (The evaluation tables were omitted because of the limitation of space.)*

## 3.3. DISCUSSION

Both systems successfully mentioned the building parts in their textual responses, reaffirming that AI systems can identify and process text and images especially when these are provided in a structured, step-by-step manner. When generating models for individual parts, Claude 3.5 Sonnet generally outperformed GPT-4o as it accurately recreates detailed elements in correct proportions, such as the four identical Entrance porticos in the Villa Rotonda. This suggests that Claude 3.5 may have a stronger capability in handling repetitive and symmetrical architectural features, which are typical characteristics in Palladio's classical architectural language. On the other hand, GPT-4o displayed better performance in recreating spatial relationships between parts, particularly in the case of Palazzo Porto. We hypothesize that this may be due to GPT-4o's ability to better process linear spatial arrangements. Both AI systems demonstrated strengths in early-stage modelling (Steps 1–3) as they effectively interpret and generate simpler geometries. The fact that the AI systems show better performance when dealing with individual component parts suggests that current AI models generally perform better with less complex tasks. In the case of 3D modelling, this means simpler geometrical parts as opposed to complex wholes that have not been sufficiently analysed or decomposed.

There are a number of ways to improve on the methodology presented here. In particular, one may further decompose complex architectural parts into simpler describable geometric shapes, integrate real-time feedback, test a different organisation of the input materials on the drawing sheets, enhance training data specificity or employ few-shot fine-tuning (Wei et al., 2023).

## 4. Conclusion

This study demonstrates that off-the-shelf Generative AI systems have the potential to augment early-stage 3D CAD modelling. It contributes in this way a benchmark for understanding the capabilities and limitations of AI systems in architectural design. The persistent difficulty in accurately representing spatial relationships and proportions suggests that current off-the-shelf Generative AI technologies lack advanced spatial reasoning and perception. This limitation impacts their ability to fully integrate geometric parts into a cohesive whole, a central aspect of architectural design synthesis and of visual-spatial perception in general (Haridis, 2020; Haridis & Pappa, 2021). Design-inspired tasks inherent in the architectural design process – interpreting abstract drawings and integrating multiple component parts into a coherent architectural whole – remain sophisticated tasks that current AI models are only beginning to approximate.

The findings show, however, that language-based Generative AI systems can still act as collaborative technical assistants for purposes of 3D CAD modelling. Further work is required both to enhance their technical capabilities and to create more intuitive, user-friendly interfaces. Such advancements would benefit architects and designers who need to rapidly test ideas and other individuals in related fields who may lack extensive technical training. At the moment, the necessity for advanced visual-spatial reasoning and an understanding of architectural principles underscores the ongoing importance of specialized training in the field. As AI systems continue to evolve, the architectural design profession will likely see a blend of human and AI-enabled expertise.